\documentclass[letterpaper, 10 pt, conference]{ieeeconf}
\IEEEoverridecommandlockouts                              

\usepackage{graphicx}
\usepackage{amsmath, amsfonts, mathrsfs}
\usepackage{caption, subcaption}
\usepackage{multirow, array, booktabs, tabularx}
\usepackage{hyperref}
\usepackage{xcolor, soul} 
\usepackage{adjustbox}
\usepackage{placeins}
\usepackage{longtable}
\let\labelindent\relax
\usepackage{enumitem}

\usepackage{cite}
\renewcommand{\arraystretch}{1.2}
\begin{document}


\title{\LARGE \bf Gait-Conditioned Reinforcement Learning with Multi-Phase Curriculum for Humanoid Locomotion}

\author{
Tianhu Peng, Lingfan Bao and Chengxu Zhou\textsuperscript{*}%
\thanks{This work was partially supported by the Royal Society [grant number RG\textbackslash R2\textbackslash232409] and the Advanced Research and Invention Agency [grant number SMRB-SE01-P06].}%
\thanks{The authors are with the Department of Computer Science, University College London, UK.}%
\thanks{\textsuperscript{*}Corresponding author, {\tt\small chengxu.zhou@ucl.ac.uk}}%
}
\bstctlcite{IEEEexample:BSTcontrol}

\maketitle

\begin{abstract}
We present a unified gait-conditioned reinforcement learning framework that enables humanoid robots to perform standing, walking, running, and smooth transitions within a single recurrent policy. A compact reward routing mechanism dynamically activates gait-specific objectives based on a one-hot gait ID, mitigating reward interference and supporting stable multi-gait learning. Human-inspired reward terms promote biomechanically natural motions, such as straight-knee stance and coordinated arm-leg swing, without requiring motion capture data. A structured curriculum progressively introduces gait complexity and expands command space over multiple phases. In simulation, the policy successfully achieves robust standing, walking, running, and gait transitions. On the real Unitree G1 humanoid, we validate standing, walking, and walk-to-stand transitions, demonstrating stable and coordinated locomotion. This work provides a scalable, reference-free solution toward versatile and naturalistic humanoid control across diverse modes and environments.
\end{abstract}

\section{Introduction}

Developing natural and efficient locomotion strategies for humanoid robots remains a core challenge in robotics. Unlike manipulators operating in structured environments, humanoids must adapt to dynamic settings where balance, adaptability, and smooth transitions are essential for real-world deployment. Human locomotion exhibits key biomechanical traits—straight-knee support, coordinated anti-phase arm swings, and smooth heel-to-toe contact—which collectively enhance balance, reduce energy expenditure, and regulate angular momentum~\cite{collins2009dynamic, pontzer2009control, herr2008armSwing, umberger2008effects}.

\begin{figure}[t]
    \centering
    \includegraphics[trim=4cm 0.8cm 6.5cm 0.3cm, clip, width=0.95\linewidth]{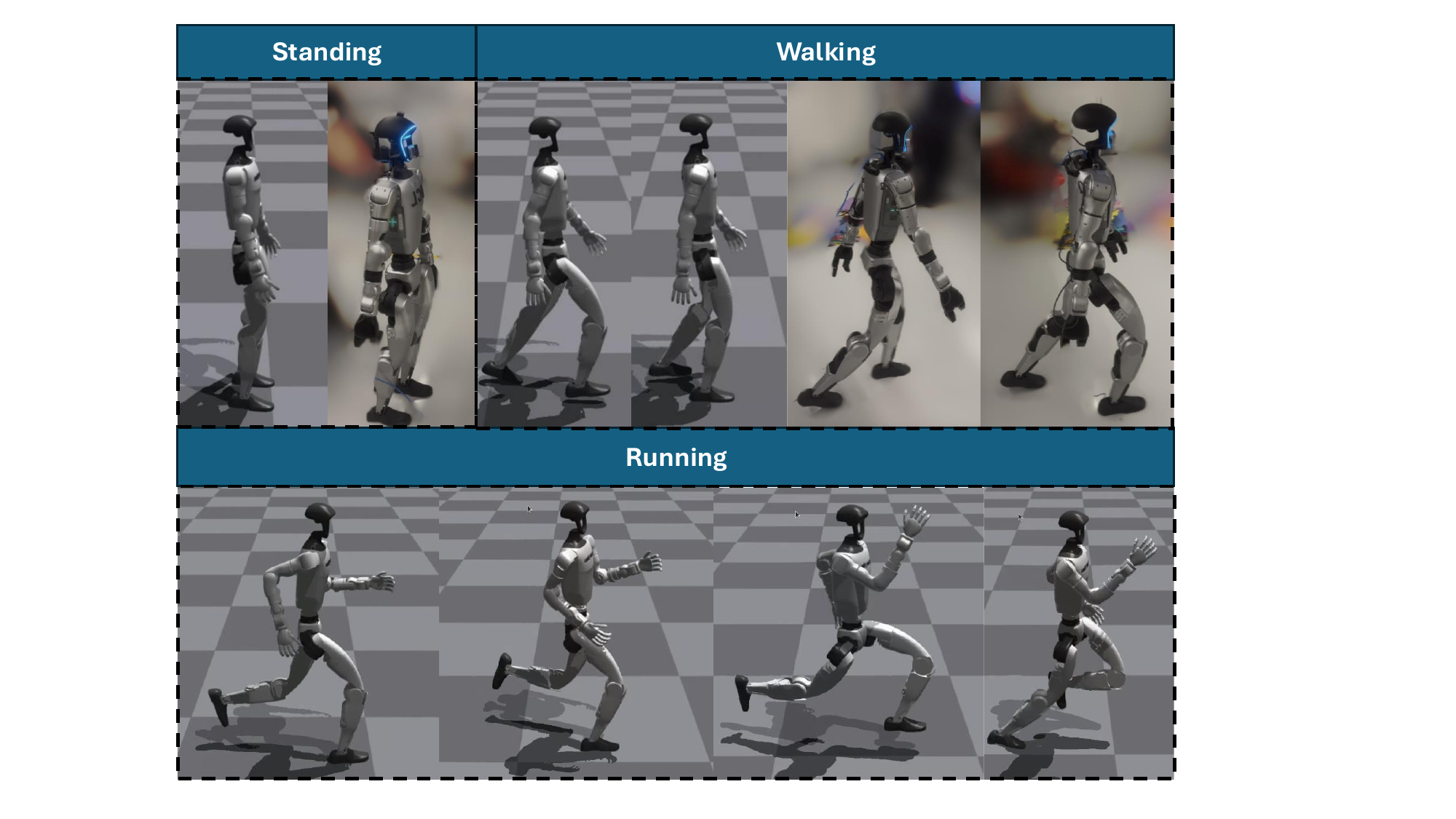}
    \caption{Human-like multi-gait locomotion on the Unitree G1 humanoid, including standing, walking, and running. The learned reference-free policy exhibits straight-knee support, coordinated arm-leg motion, and natural transitions without MoCap references.}
    \label{fig:Running}
\end{figure}

These traits are not merely stylistic but serve fundamental functions. Rhythmic arm swing, for instance, helps cancel leg-induced angular momentum, improving trunk stability and reducing metabolic cost. While recent reinforcement learning (RL) work shows emergent anti-phase swing as a side effect of energy minimization~\cite{radosavovic2024real}, such methods rarely enforce angular momentum control directly. In contrast, we propose a biomechanics-inspired reward that explicitly penalizes residual angular momentum and promotes phase-symmetric arm-leg coordination—yielding stable and efficient gaits without relying on trajectory references.

Recent RL advances have enabled agile locomotion for legged robots~\cite{2025_DRL_Bipedal_review}. However, reference-based methods such as Adversarial Motion Priors (AMP)~\cite{peng2021amp} require large-scale motion capture datasets and often struggle with morphology mismatch between human demonstrators and robot platforms. Moreover, their reliance on implicit imitation objectives limits interpretability, and prevents flexible reward design or task-specific modulation.

These limitations are amplified when integrating multiple gaits/skills into one controller. Prior approaches rely on multi-policy distillation~\cite{zhuang2023robot}, mixture-of-experts~\cite{luo2023perpetual}, or skill fusion~\cite{peng2019mcp}, which typically require expert pretraining, switching mechanisms, and careful reward coordination across sub-policies. These architectures increase system complexity and training cost, and often suffer from interference between skill domains during deployment.

We propose a simpler alternative: a unified, gait-conditioned RL framework where a single recurrent policy learns standing, walking, running, and smooth transitions. A gait-conditioned reward routing mechanism activates gait-specific rewards based on a compact gait ID in the observation, mitigating interference and supporting stable multi-gait training.

To encourage human-like motion, we incorporate biomechanically grounded reward terms that promote straight-knee support, arm-leg coordination, minimal foot drag, and push-off dynamics. Our structured multi-phase curriculum enables progressive skill acquisition in simulation and robust deployment on hardware. This curriculum is inspired by biological motor development and bio-inspired gait adaptation studies~\cite{humphreys2024bio}. A visual overview of the learned behaviors is shown in Fig.~\ref{fig:Running}. In summary, our contributions are:
\begin{itemize}
    \item A unified, reference-free RL framework for standing, walking, running, and transitions in a single recurrent policy.
    \item A gait-conditioned reward routing scheme that mitigates reward interference.
    \item Biomechanically grounded reward shaping for efficient, natural locomotion without MoCap.
    \item A progressive multi-phase curriculum for skill expansion and stable training.
\end{itemize}

\section{Related Work}

Learning natural and efficient locomotion has been a longstanding goal in RL for legged robots, especially in bipedal settings where stability and versatility are more demanding~\cite{2025_DRL_Bipedal_review}.
 Existing methods can be broadly categorized into reference-based learning, reference-free approaches, modular multi-skill frameworks, and multi-behavior learning via reward or value function decoupling.

\subsection{Reference-Based Locomotion Learning}

Early successes in humanoid locomotion were largely achieved through reference-based approaches, where policies imitate curated motion capture (MoCap) data.
A seminal example is AMP~\cite{peng2021amp}, which introduced an adversarial framework to produce human-like motions without explicit trajectory tracking. In multi-gait settings, AMP uses several motion clips (e.g., trotting, pacing) with distinct root velocity and angular velocity profiles, switching between them based on commanded velocity. A velocity-tracking reward aligns the policy’s motion with the reference clip’s kinematics, effectively mapping different velocities to specific MoCap segments.
AMP has been extended to quadrupeds~\cite{escontrela2022adversarial}, humanoid whole-body control~\cite{ampforhumanoid}, and cross-morphology transfer such as enabling quadrupeds to walk bipedally~\cite{peng2024learning}.

Beyond MoCap imitation, other reference-based methods leverage analytically generated trajectories or pre-defined gait libraries. Residual learning~\cite{2020_Xie_firstsim2real_,2021_siekmann_sim2real_nonreference_perodicreward_DRL_e2e_LSTM_PPO_cassie} refines a nominal reference by predicting additive action offsets, enhancing adaptability while retaining its structure. Guided learning directly tracks analytical references such as Hybrid Zero Dynamics (HZD)\cite{2021_UCB_hybridrobotics_sim2real_referencebased_HZD_gaitlibrary_e2epolicy_drl_Cassie_lowpassfilter} or Central Pattern Generator (CPG) templates\cite{hwangbo2019learning}, embedding rhythmic priors into training.

While effective, these methods depend on high-quality MoCap or analytical trajectories and often fail to generalize beyond their distribution. Morphological mismatches—differences in limb proportions, joint limits, or mass distribution—can degrade performance, and controllers trained on narrow gait libraries may struggle with unseen commands or disturbances.

\subsection{Reference-Free RL}

Reference-free RL dispenses with predefined trajectories, instead optimizing handcrafted rewards to encourage energy-efficient, symmetric, and periodic gaits~\cite{2023_vanmarum_visionDRL_studentteacher_irregularterrain_PPO_periodicrewardfunction}. Early work~\cite{2018_wenhao_yu_DRL_withoutpredefine_symmetrygait_PPO_jointanglePD_} demonstrated bipedal gait learning via symmetry and curriculum shaping; later studies targeted stepping stone navigation~\cite{2020_zhaoming_drl_steppingstones_PPOwithactorcritic_referencefree_simulation} and achieved sim-to-real transfer on Cassie with periodic rewards~\cite{2021_siekmann_sim2real_nonreference_perodicreward_DRL_e2e_LSTM_PPO_cassie}, extended to blind stair climbing~\cite{2021_siekmann_blind_DRL_stair} and vision-guided footstep placement~\cite{2022_sim2real_footstepconstraint_OSUDRL_specifytouchdownlocation_actorcritic_PPO_LSTM_plannar_transisionmodel_model-predictiveplanning_CNN_predictnexttdlocation,2022_OSUDRL_steppingstone_referencefree_predictionfeasiblefootsteps_camera_benchmark}. Heightmap-based perception further broadened terrain generalization~\cite{2023_duan_OSUDRL_heightmap_visionbased_hybrid}.  

While enabling greater generalization and design freedom, reference-free methods require careful reward tuning, converge slowly, and struggle with complex maneuvers such as agile transitions or jumping.

\subsection{Multi-Skill and Modular Architectures}

Multi-skill RL aims to enable a single agent to execute diverse behaviors.  
Supervised strategies such as policy distillation~\cite{rusu2015policy} and DAgger~\cite{ross2011reduction} aggregate expert demonstrations to train a unified policy.  
For example, Han et al.~\cite{han2024lifelike} distill multiple task-specific controllers into a single terrain-adaptive locomotion policy, while Zhuang et al.~\cite{zhuang2023robot} use DAgger to learn robust parkour behaviors from expert rollouts.  
These methods achieve strong performance but depend on extensive expert data and carefully managed curricula, limiting scalability to unseen behaviors or larger gait sets.

Modular approaches manage multiple skills via decomposed control structures.  
Mixture-of-Experts (MoE) controllers~\cite{luo2023perpetual} dynamically select sub-policies, as in Perpetual Humanoid Control, which achieved real-time coordination of diverse human-like behaviors.  
Multiplicative Compositional Policies (MCP)~\cite{peng2019mcp} blend low-level primitives via multiplicative composition for flexible skill recombination.  
While modular frameworks offer high flexibility, they often incur significant architectural complexity and require skill-specific supervision or pretraining.

\subsection{Reward Routing and Value Function Decoupling}

A key challenge in multi-behavior RL is mitigating reward interference across heterogeneous skills. Approaches such as MultiCritic Actor Learning~\cite{mysore2022multi} address this by maintaining a single actor with task-specific critic heads, improving stability. In locomotion, CPG- or phase-conditioned policies~\cite{hwangbo2019learning, rudin2022learning} exploit rhythmic motion priors to stabilize multi-gait control.

While prior work like DeepMimic~\cite{peng2018deepmimic} incorporates one-hot skill identifiers into the observation to differentiate motions, these are used solely for policy conditioning without altering the reward structure. In contrast, our framework employs gait IDs both for conditioning and for dynamically routing gait-specific rewards, enabling structured multi-gait learning within a unified policy.

Despite these advances, achieving smooth and efficient gait transitions without explicit experts, motion references, or hierarchical planners remains an open challenge.

\subsection{Our Approach}

Unlike AMP-based approaches~\cite{peng2018deepmimic, peng2021amp} that require large-scale motion capture datasets and often map commanded velocities to pre-recorded gait clips via velocity tracking rewards, our method operates entirely without motion references. Instead of using MoCap-derived velocity profiles to drive gait switching, we define each gait—standing, walking, running, and transitions—directly through human-inspired biomechanical reward shaping, enabling the policy to discover natural transitions without clip-based supervision.
In contrast to multi-policy or distillation frameworks~\cite{rusu2015policy, ross2011reduction} that train and merge multiple specialized experts, we employ a single recurrent policy with a compact gait-ID encoding, allowing all gaits to be learned end-to-end within one unified architecture. This design eliminates expert-switching logic, reduces model complexity, and mitigates reward interference across gaits via a simple yet effective gait-conditioned reward routing mechanism. Compared to prior modular, distillation-based, or reference-driven methods, our framework emphasizes minimal architectural overhead, unified end-to-end training, and strong generalization across diverse locomotion behaviors, offering a scalable pathway toward robust and naturalistic humanoid locomotion control.

\section{Methodology}

\subsection{Problem Setup}

We address the problem of learning unified multi-gait control for humanoid locomotion, encompassing standing, walking, running, and transitions between these modes. The task is formulated as a partially observable Markov decision process (POMDP), defined by a tuple \((s_t, a_t, P, r_t, p_0, \gamma)\), where \(s_t\) represents the latent system state at time \(t\), \(a_t\) is the action executed by the agent, \(P\) denotes the transition dynamics, \(r_t\) is the reward function, \(p_0\) is the initial state distribution, and \(\gamma\) is the discount factor. The objective is to learn a policy \(\pi_\theta\) that maximizes the expected cumulative discounted return:
\begin{equation}
J(\theta) = \mathbb{E}_{\pi_\theta} \left[\sum_{t=0}^{\infty} \gamma^t r_t \right]
\end{equation}

\subsection{Policy Architecture}

Our policy adopts an asymmetric actor–critic architecture~\cite{andrychowicz2020learning}, with both networks implemented as Long Short-Term Memory (LSTM)~\cite{hochreiter1997lstm} layers followed by Multi-Layer Perceptrons (MLPs). The actor, designed for deployment, receives only proprioceptive observations, while the critic is given additional privileged information during training to improve value estimation.

The LSTM captures temporal dependencies across gait cycles, aiding balance recovery, handling delayed contact effects, and enabling smooth multi-gait transitions under partial observability. Actor inputs include base angular velocity and gravity orientation in the local frame, commanded velocities, joint position offsets from a nominal pose, joint velocities, the previous action, a sine–cosine gait phase encoding, and a one-hot gait ID (e.g., standing, walking, running, transition). The network outputs a 23-dimensional joint position offset vector, tracked by a joint-level proportional–derivative (PD) controller with fixed gains.

The critic augments the actor’s inputs with privileged features such as foot contact states, ground reaction forces, contact friction/restitution, binary contact indicators at hip/knee joints, and externally applied disturbances with application points. This improves stability and accuracy in value estimation, even in noisy or partially observable settings.

\begin{figure*}
    \centering
    \includegraphics[trim = 17mm 40mm 68mm 35mm, clip, width=1.0\linewidth]{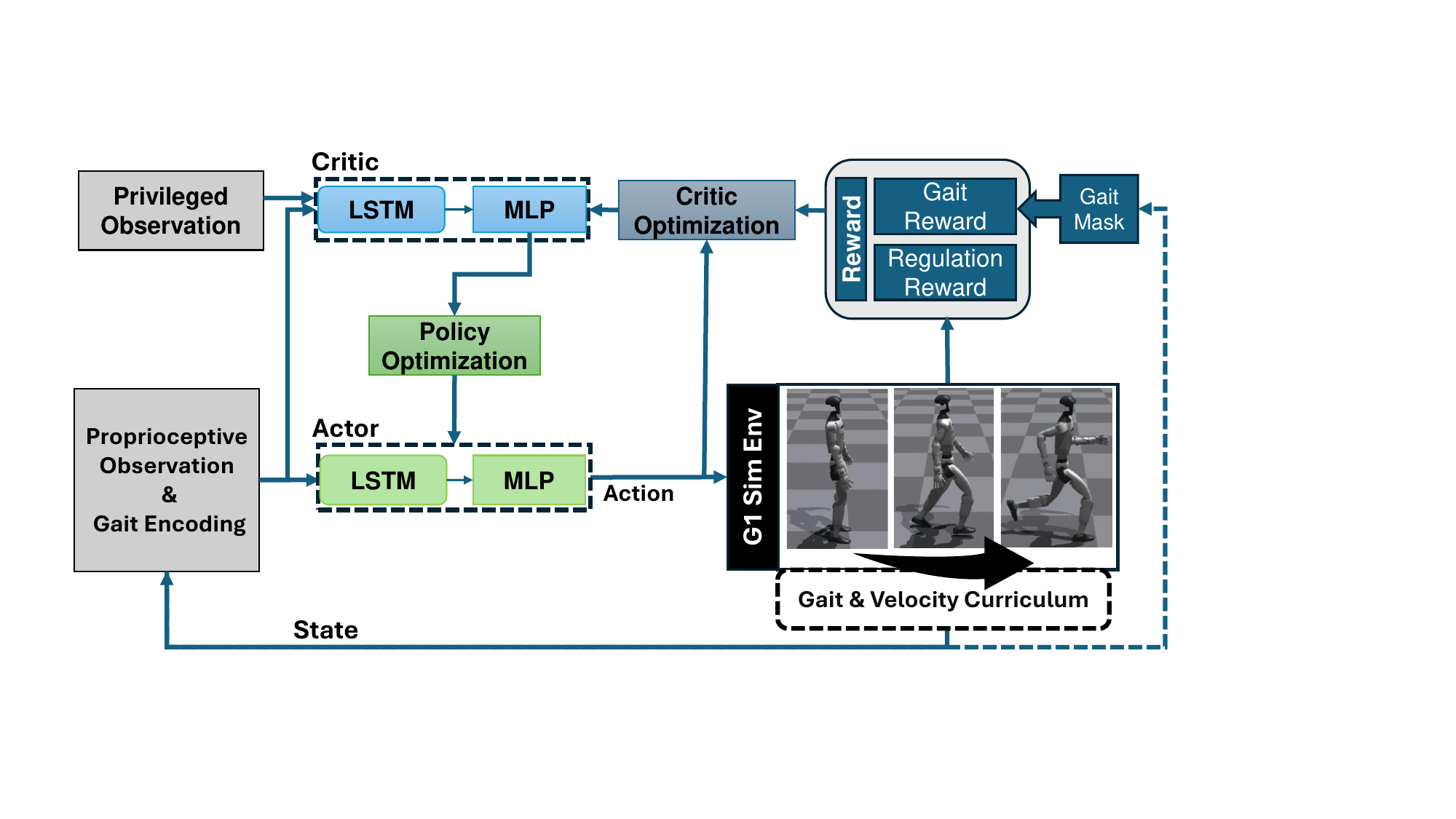}
    \caption{Overview of our proposed gait-conditioned RL framework. A recurrent actor receives proprioceptive states with gait encoding. Gait-conditioned reward masks route mode-specific rewards, enabling multi-gait learning.}
    \label{fig:framework}
\end{figure*}

An overview of the proposed gait-conditioned RL framework—highlighting the asymmetric architecture, gait-conditioned reward routing, and multi-phase curriculum—is shown in Fig.~\ref{fig:framework}, illustrating how gait mode information conditions both actor and critic observations, from sensory input through network processing, reward computation, and final action selection.



\subsection{Gait-Conditioned Reward Routing}

To enable multi-gait locomotion within a unified policy, we adopt a gait-conditioned reward routing mechanism. At each timestep, a gait mode ID is inferred based on the commanded velocity and the robot's dynamic state. This ID is encoded as a one-hot vector and appended to the policy's observation, conditioning the actor-critic network on the intended gait mode.

\begin{figure*}
    \centering
    \includegraphics[trim = 66mm 26mm 74mm 21mm, clip, width=0.7\linewidth]{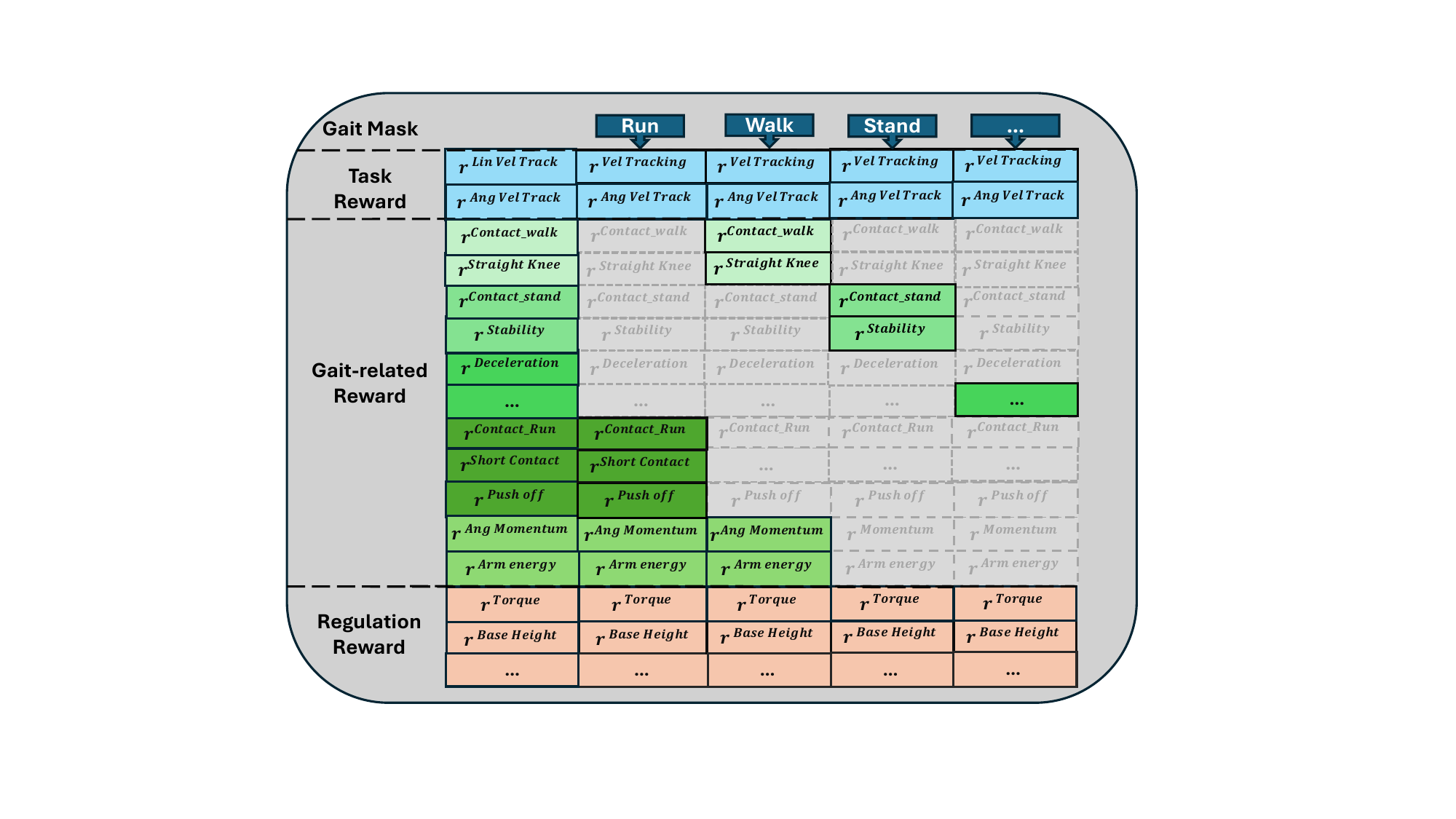}
    \caption{Illustration of gait-conditioned reward routing during three example modes: Run, Walk, and Stand. 
    The unified reward vector includes task-level and regulation rewards (shared across all gaits), as well as gait-specific terms (e.g., contact, push-off, or stability). 
    A gait mask is used to activate only the relevant terms at each timestep. 
    Ellipses (\textellipsis) indicate omitted modes (e.g., transitions) or rewards.}
    \label{fig:gait-mask-reward}
\end{figure*}

During training, a gait mask is applied to selectively activate mode-specific reward terms according to the current gait. As shown in Fig.~\ref{fig:gait-mask-reward}, the complete reward vector includes shared components (e.g., task-level tracking and regulation rewards) as well as gait-related terms (e.g., contact patterns or push-off dynamics) specific to walking, running, or standing. The gait mask filters the reward vector such that only the relevant subset contributes to the learning signal. For example, running activates push-off and short contact duration rewards; walking enables swing height and foot symmetry terms; standing focuses on upright posture and stillness bonuses while disabling locomotion-specific terms.

This design ensures that each locomotion behavior—standing, walking, running, and their transitions—is optimized independently while coexisting within a single recurrent policy. It effectively mitigates reward conflicts, stabilizes multi-gait learning, and enables clean credit assignment across gait modes.

\subsection{Curriculum Learning Strategy}

Inspired by the natural progression of human motor development—from static balance to walking and eventually running—we design a biologically grounded curriculum to enable stable acquisition of locomotion skills.

Training all gaits simultaneously often leads to reward conflict and unstable exploration. To address this, we adopt a multi-phase curriculum that gradually introduces gait modes, broadens the commanded velocity range, and incrementally activates coordination mechanisms.

The curriculum progresses along three dimensions:
\begin{itemize}
\item Gait complexity: Starting from walking, we progressively add standing, running, and transitional modes.
\item Commanded velocity: The target velocity range expands from low-speed walking to high-speed running (up to $4.0$~m/s).
\item Coordination mechanisms: Reward routing, gait-aware masking, and smooth transition blending are gradually enabled.
\end{itemize}

We train the policy in three phases:

\begin{itemize}
    \item \textbf{Phase 1} --- Walking only: Only \texttt{Walk} (ID = 1) is enabled. The gait ID is fixed to \texttt{Walk} at all timesteps; other modes are disabled, and the reward mask activates walking-specific terms only.  This stage develops periodic locomotion with stable contact, adequate foot clearance, and extended knees.

    \item \textbf{Phase 2} --- Standing and Walk to Stand (W2S): We introduce gait-ID switching and corresponding rewards. When the commanded speed norm satisfies $\|\mathbf{v}_c\| < 0.1$\,m/s, the system enters \texttt{W2S} (ID = 2). If low speed and double support persist for a manually specified $1.5$\,s, it switches to \texttt{Stand} (ID = 0). This hysteresis prevents premature switching.

    \item \textbf{Phase 3} --- Running and Run-to-Walk (R2W): \texttt{Run} (ID = 3) is enabled using the Froude criterion $Fr = v^2/(g\,l)$ with $Fr>0.5$. Upon speed reduction below this regime, the system enters \texttt{R2W} (ID = 4) until a manually specified $2.5$\,s stable duration completes, ensuring smooth velocity decay and contact reshaping. Running- and transition-specific rewards are activated, and all five modes (\texttt{Stand}, \texttt{Walk}, \texttt{W2S}, \texttt{Run}, \texttt{R2W}) are co-trained with full reward routing.
\end{itemize}

This progressive curriculum improves training stability, reduces reward interference, and facilitates generalization across diverse locomotion contexts.

\subsection{Human-Inspired Gait Design and Reward Shaping}

While RL can produce stable locomotion behaviors, the resulting motions often appear overly crouched or energetically suboptimal. To encourage more natural and efficient gait patterns, we incorporate biomechanical principles observed in human locomotion directly into the reward design.

Specifically, our approach draws on the following insights:
\begin{itemize}
    \item Straight-knee support improves force transmission and reduces muscular effort during stance phases~\cite{zajac2002biomechanics}.
    
    \item Anti-phase arm-leg coordination mitigates angular momentum buildup and stabilizes the upper body~\cite{collins2009dynamic, pontzer2009control}.
    
    \item Gait transitions are gradual and biomechanically necessary—humans do not switch abruptly between running, walking, or standing, but instead adopt intermediate steps to reduce momentum and maintain balance. This motivates our inclusion of explicit transition gaits (e.g., walk-to-stand, run-to-walk), which align with observed human motor strategies~\cite{minetti2000transition, hubel2013transition}.
\end{itemize}

These principles are encoded through both gait-specific and shared reward components. Table~\ref{tab:reward-gaits} summarizes representative terms used to characterize and stabilize each locomotion mode. While the full reward set includes additional objectives (e.g., symmetry, torque minimization), we omit them here for brevity.

\begin{table}
\setlength{\abovecaptionskip}{4pt}
\setlength{\belowcaptionskip}{2pt}
\caption{Representative Human-Inspired Reward Terms by Gait Mode}
\label{tab:reward-gaits}
\renewcommand{\arraystretch}{1.05}
\setlength{\tabcolsep}{2.5pt}
\centering
\begin{adjustbox}{max width=\linewidth}
\begin{tabularx}{\linewidth}{p{13mm}|p{18mm}|X}
\toprule
\textbf{Gait} & \textbf{Reward Term} & \textbf{Description} \\
\cmidrule(lr){1-3}
\multirow{3}{*}{Walking} 
& Contact Pattern & Encourage phase-aligned foot contacts based on cyclic gait timing. \\
\cmidrule(lr){2-3}
& Foot Clearance & Promote sufficient foot lift during swing to avoid dragging. \\
\cmidrule(lr){2-3}
& Straight Knee & Encourage extended knee during stance to improve support efficiency. \\
\cmidrule(lr){1-3}
\multirow{3}{*}{Running} 
& Contact Pattern & Encourage alternating single-leg contact and flight phases. \\
\cmidrule(lr){2-3}
& Push-Off Dynamics & Reward strong vertical and forward velocity during push-off. \\
\cmidrule(lr){2-3}
& Short Contact & Penalize prolonged stance to promote dynamic running. \\
\cmidrule(lr){1-3}
\multirow{2}{*}{Standing} 
& Contact Pattern & Encourage consistent double-foot support for static balance. \\
\cmidrule(lr){2-3}
& Base Stability & Penalize base and joint motion to maintain upright posture. \\
\cmidrule(lr){1-3}
\multirow{2}{*}{Transition} 
& Contact Pattern & Promote correct contact phasing during gait switching. \\
\cmidrule(lr){2-3}
& Smooth Deceleration & Encourage gradual reduction in velocity to settle into stance. \\
\bottomrule
\end{tabularx}
\end{adjustbox}
\vspace{-0.4em}
\caption*{\footnotesize 
Key gait-specific rewards used to stabilize and differentiate locomotion behaviors. 
Additional terms such as symmetry and coordination are used but omitted for brevity.}
\end{table}

\noindent
\subsubsection{Angular Momentum Compensation via Arm-Leg Coordination}

Human arm swing plays a critical role in balancing the whole-body angular momentum during walking. Biomechanical studies~\cite{collins2009dynamic, pontzer2009control, herr2008armSwing} have shown that arm motion primarily serves to counteract leg-induced angular momentum—especially in the yaw direction—thereby enhancing trunk stability and reducing energy expenditure.

Inspired by this, we design a reward to promote human-like, dynamically balanced arm swing without relying on trajectory references. The total centroidal angular momentum $\mathbf{L}_{\text{total}}$ is computed via the standard decomposition~\cite{lee2012centroidal}

To encourage coordinated arm-leg motion, we define the angular momentum reward as:
\begin{align}
R_{\text{momentum}} =\; & - \left( L_{\text{total},z}^2 \right) 
 - 0.4\, \left( L_{\text{la},z} - L_{\text{ra},z} \right)^2
\end{align}
where $L_{\text{la}}, L_{\text{ra}} \in \mathbb{R}^3$ denote the angular momentum vectors of the left and right arms, and the subscripts $x$ and $z$ refer to pitch and yaw axes.

This reward encourages:
\begin{itemize}
    \item \textit{Whole-body momentum minimization:} Reduces residual angular momentum, particularly in the yaw direction;
    \item \textit{Anti-phase yaw swing:} Promotes alternating arm motion to counteract leg-induced rotation.
\end{itemize}

\subsection{Training Platform}

Training is performed in Isaac Gym \cite{rudin2022learning}, a GPU-accelerated physics simulator supporting parallel simulation of 1000 humanoid environments with NVIDIA RTX 4090 GPU. Policy optimization employs Proximal Policy Optimization (PPO) \cite{schulman2017proximal}, with recurrent neural networks for both actor and critic, ensuring high sample efficiency and robust policy performance under partial observability.

\subsection{Domain Randomization}

To improve policy robustness and facilitate sim-to-real transfer, we apply domain randomization during training. Several physical parameters are perturbed, including ground friction coefficient, robot base mass, and joint control properties. Additionally, external disturbances are introduced by applying randomized lateral pushes to the robot's base velocity at fixed intervals. 

These perturbations expose the policy to a wide range of dynamics and environmental variations, encouraging generalizable locomotion behaviors that remain stable under uncertainty.

\section{Experiment and Results}

We evaluate our proposed gait-conditioned RL framework both in simulation (Isaac Gym) and real robot . The policy is trained to perform standing, walking, running, and smooth transitions between these modes using a single recurrent controller. Our experiments demonstrate that the learned policy achieves robust gait switching across varying commanded velocities, while exhibiting human-like motion and energy-efficient, coordinated behaviors—without relying on motion capture references or hierarchical planners.

\subsection{Simulation Results}

\subsubsection{Gait Switching and Velocity Tracking}

\begin{figure} \centering
\includegraphics[width=0.95\linewidth]{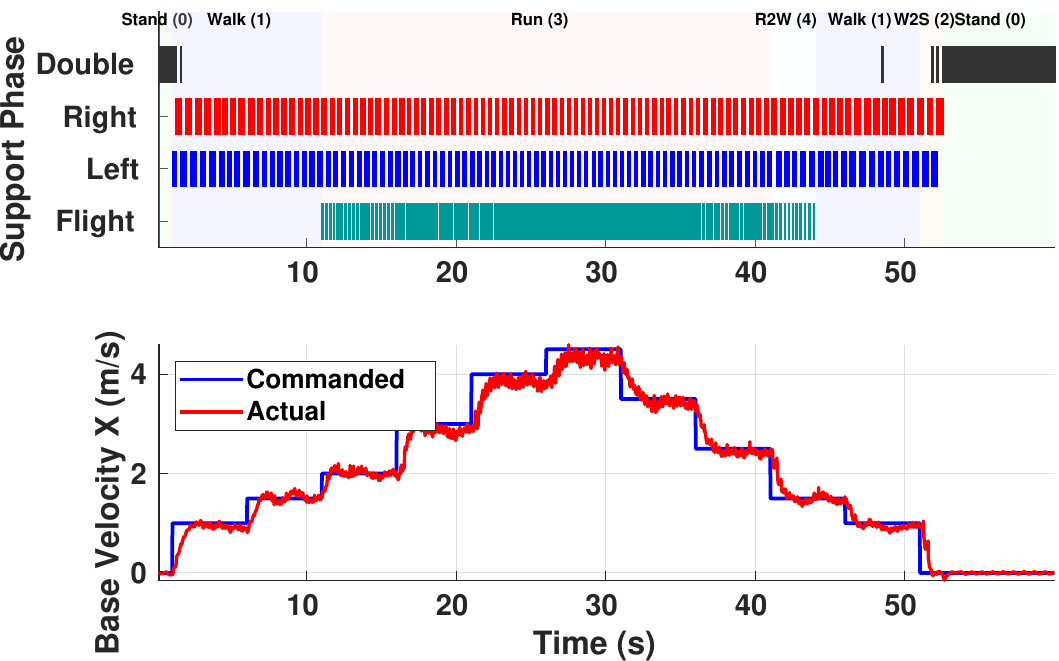}
\caption{
Top: Support contact classification with background gait labels including Stand, Walk, Run, W2S, and R2W. 
Bottom: Commanded vs. actual forward base velocity. Gait transitions are smooth, dynamically consistent, and reflect internal coordination.}
\label{fig:gait_phase_velocity}
\end{figure}

To evaluate the controller's adaptability to varying locomotion demands, we command a velocity profile that ramps up and down between $0$ and $3$ m/s. Fig.~\ref{fig:gait_phase_velocity} (bottom) shows that the actual forward velocity closely follows the commanded profile across a range of gaits. The top subplot visualizes contact phases along with background gait mode labels.

The policy exhibits stable transitions between standing (green), walking (blue), running (orange), and transitional gaits: W2S (purple) and R2W (red). These transitions occur smoothly without abrupt contact shifts, indicating robust internal gait modulation. In particular, the policy slows down naturally from running into walking , and settles into a balanced standing posture without external resets or mode switches.

\subsubsection{Angular Momentum Coordination}

\begin{figure} \centering
\includegraphics[width=0.95\linewidth]{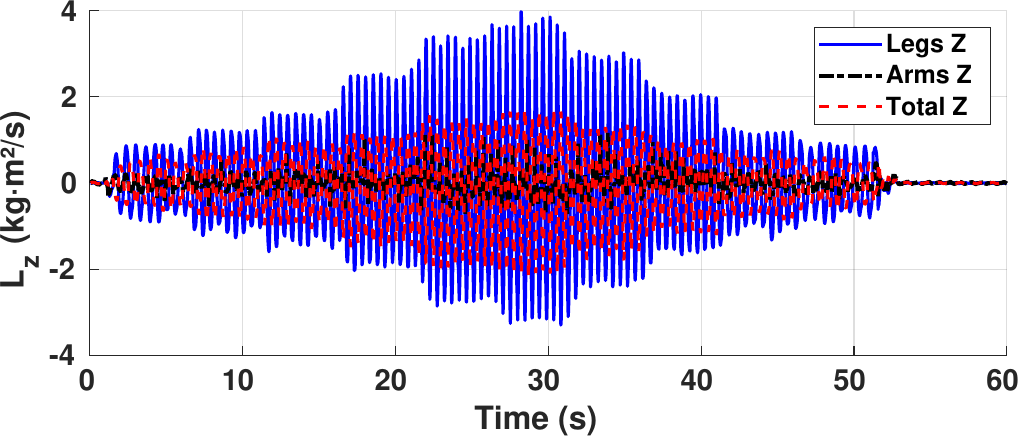}
\caption{
Z-axis angular momentum of legs (blue solid), arms (red dashed), and total (black dash-dot). 
The policy achieves momentum coordination across standing, walking, running, and transitional gaits (W2S, R2W), with effective cancellation between arm and leg segments.}
\label{fig:momentum}
\end{figure}

Fig.~\ref{fig:momentum} shows the Z-axis angular momentum of legs, arms, and the combined body over time. Throughout different gait modes—including transitions—the policy maintains low total angular momentum by coordinating the motion of arms and legs in anti-phase.

In walking and running, the arm and leg momenta are of opposite sign and similar magnitude, resulting in effective cancellation. During the R2W and W2S transitions, the coordination continues smoothly as the amplitude of each segment adapts to the underlying velocity and contact mode. This demonstrates that the learned behavior is not only dynamically stable, but also biomechanically efficient.

\vspace{0.5em}
\noindent These results confirm that our gait-conditioned policy learns to:  
(i) switch naturally between gaits;  
(ii) track commands robustly; and  
(iii) maintain biomechanical coordination.

\subsubsection{Ablations: Curriculum and Reward Routing}

We ablate two core components under identical architecture, hyper-parameters, and domain randomization:  
(i) \emph{No Curriculum}: train all five gait modes (Standing, Walking, Running, W2S and R2W) jointly from scratch;  
(ii) \emph{No Routing}: retain curriculum and gait ID in the observation, but disable reward routing so that all gait-specific terms (Table~\ref{tab:reward-gaits}) remain active simultaneously (mask removed).  
All settings are trained for an equal budget and evaluated on the same validation episodes, with results summarized in Table~\ref{tab:ablation-core}.

\begin{table}[h]
\centering
\caption{Ablations under equal training budget. Len: mean episode length; Ret: mean return.}
\label{tab:ablation-core}
\setlength{\tabcolsep}{6pt}
\renewcommand{\arraystretch}{1.05}
\begin{tabular}{lcc}
\toprule
Setting & Len. $\uparrow$ & Ret. $\uparrow$ \\
\midrule
\textbf{Ours} (Curriculum + Routing) & $972.6$ & $89.3$ \\
No Curriculum (5 gait modes)         & $31.26$ & $2.45$ \\
No Reward Routing (all terms active) & $11.59$ & $0.00$ \\
\bottomrule
\end{tabular}
\end{table}

Removing the curriculum forces the policy to learn all gait modes simultaneously from scratch, which greatly increases the optimization difficulty and prevents convergence within the same budget (episode length $31.26$ vs.\ $972.6$, return $2.45$ vs.\ $89.3$).  
Disabling reward routing, even with curriculum, activates mutually conflicting gait-specific rewards at every timestep (e.g., stillness from standing vs.\ forward velocity from walking/running), causing unstable gradient updates and rapid performance collapse (episode length $11.59$, return $0.00$).

\subsection{Real-World Transfer}

To evaluate the transferability of our policy from simulation to physical hardware, we deploy the learned controller on a real Unitree G1 humanoid robot. The deployed policy is directly transferred without any additional fine-tuning, sim-to-real adaptation, or dynamics randomization beyond what was already applied during training.

Our system successfully demonstrates stable standing, smooth walk-to-stand transitions, and  walking on real hardware. These behaviors remain coherent and robust under moderate command perturbations and exhibit natural whole-body coordination. In particular, the arm-leg coordination and double-foot support in standing mode transfer well, indicating the effectiveness of our human-inspired reward shaping and gait-conditioned training framework.

\begin{figure*}
    \centering
    \includegraphics[width=\linewidth]{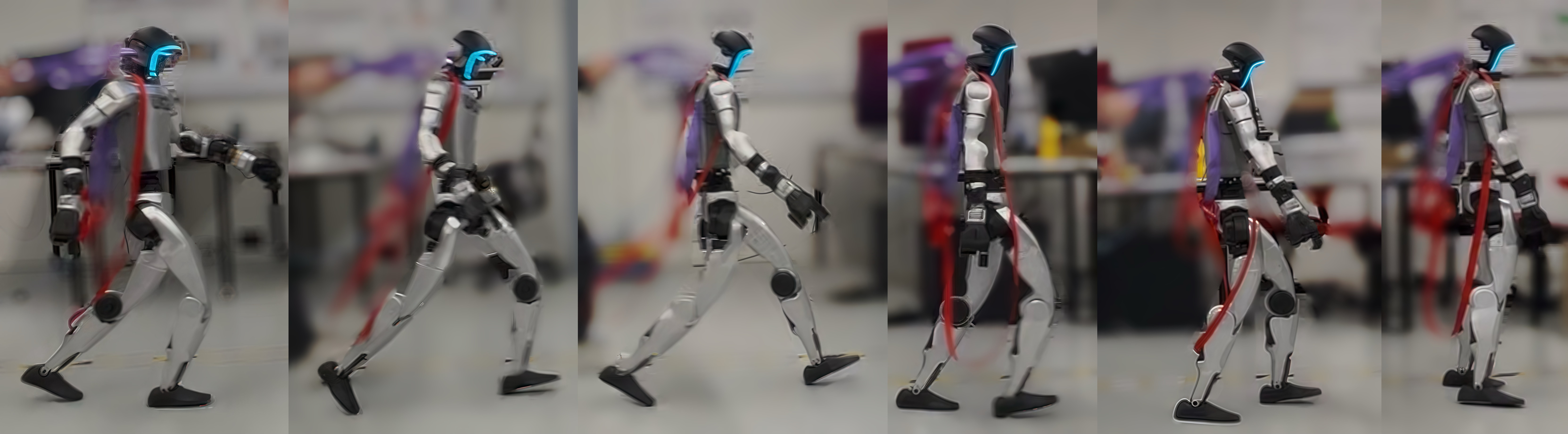}
    \caption{Real-world deployment of our learned policy on the Unitree G1 humanoid. The robot demonstrates smooth transitions between walking and standing, with natural limb coordination and upright posture—transferred directly from simulation without any fine-tuning.}
    \label{fig:real-walk}
\end{figure*}

Figure~\ref{fig:real-walk} shows snapshots of the deployed policy performing walking on the physical Unitree G1. The robot maintains biomechanically plausible postures such as extended knees and coordinated arm-leg motion, closely matching the simulated behavior.

Overall, our results validate that the proposed framework generalizes effectively across modalities and environments, with promising performance in real-world humanoid locomotion.

\section{Conclusion and Future Work}

We proposed a unified gait-conditioned RL framework for humanoid locomotion, combining gait-specific reward routing, biomechanically inspired reward shaping, and multi-phase curriculum learning. The resulting recurrent policy enables standing, walking, running, and smooth transitions, all within a single controller and without reliance on motion capture data.

Our method achieves robust and naturalistic multi-gait behaviors in physics-based simulation using a Unitree G1 humanoid. Initial real-world deployment demonstrates successful transfer of standing, walking, and walk-to-stand transitions. However, gait classification currently relies on manually defined gait IDs based on commanded velocity and Froude number, which constrains the emergence of novel gait patterns. The approach also depends on extensive manual design and tuning of numerous gait-specific reward terms, making it labor-intensive and increasingly challenging for larger or more complex gait sets. Moreover, as the number of gaits grows, the shared policy may be prone to catastrophic forgetting or behavioral drift due to overlapping objectives.

We are currently working to extend this to dynamic running behaviors, which are already stable in simulation. Future directions include scaling to more complex scenarios such as traversal of uneven or deformable terrain, vision-guided locomotion in dynamic environments, and the incorporation of task-conditioned objectives for whole-body behaviors like climbing, object interaction, and human-robot collaboration.

\bibliographystyle{IEEEtran}
\bibliography{Reference}

\begin{thebibliography}{10}
\providecommand{\url}[1]{#1}
\csname url@samestyle\endcsname
\providecommand{\newblock}{\relax}
\providecommand{\bibinfo}[2]{#2}
\providecommand{\BIBentrySTDinterwordspacing}{\spaceskip=0pt\relax}
\providecommand{\BIBentryALTinterwordstretchfactor}{4}
\providecommand{\BIBentryALTinterwordspacing}{\spaceskip=\fontdimen2\font plus
\BIBentryALTinterwordstretchfactor\fontdimen3\font minus
  \fontdimen4\font\relax}
\providecommand{\BIBforeignlanguage}[2]{{%
\expandafter\ifx\csname l@#1\endcsname\relax
\typeout{** WARNING: IEEEtran.bst: No hyphenation pattern has been}%
\typeout{** loaded for the language `#1'. Using the pattern for}%
\typeout{** the default language instead.}%
\else
\language=\csname l@#1\endcsname
\fi
#2}}
\providecommand{\BIBdecl}{\relax}
\BIBdecl

\bibitem{collins2009dynamic}
S.~H. Collins, P.~G. Adamczyk, and A.~D. Kuo, ``Dynamic arm swinging in human
  walking,'' \emph{Proc. Royal Society B: Biological Sciences}, vol. 276, no.
  1673, pp. 3679--3688, 2009.

\bibitem{pontzer2009control}
H.~Pontzer, J.~H. Holloway \emph{et~al.}, ``Control and function of arm swing
  in human walking and running,'' \emph{Journal of Experimental Biology}, vol.
  212, no.~4, pp. 523--534, 2009.

\bibitem{herr2008armSwing}
H.~Herr and M.~Popovic, ``The roles of arm swing in human walking,''
  \emph{Journal of Experimental Biology}, vol. 211, no.~4, pp. 633--640, 2008.

\bibitem{umberger2008effects}
B.~R. Umberger, ``Effects of suppressing arm swing on kinematics, kinetics, and
  energetics of human walking,'' \emph{Journal of Biomechanics}, vol.~41,
  no.~11, pp. 2575--2580, 2008.

\bibitem{radosavovic2024real}
I.~Radosavovic, A.~Desai \emph{et~al.}, ``Real-world humanoid locomotion with
  reinforcement learning,'' \emph{Science Robotics}, vol.~9, no.~87, p.
  eadi9579, 2024.

\bibitem{2025_DRL_Bipedal_review}
L.~Bao, J.~Humphreys \emph{et~al.}, ``Deep reinforcement learning for bipedal
  locomotion: A brief survey,'' \emph{arXiv preprint arXiv:2404.17070}, 2025.

\bibitem{peng2021amp}
X.~B. Peng, Z.~Ma \emph{et~al.}, ``Amp: Adversarial motion priors for stylized
  physics-based character control,'' \emph{ACM Transactions on Graphics},
  vol.~40, no.~4, pp. 1--20, 2021.

\bibitem{zhuang2023robot}
Z.~Zhuang, Z.~Fu \emph{et~al.}, ``Robot parkour learning,'' \emph{arXiv
  preprint arXiv:2309.05665}, 2023.

\bibitem{luo2023perpetual}
Z.~Luo, J.~Cao \emph{et~al.}, ``Perpetual humanoid control for real-time
  simulated avatars,'' in \emph{Proc. IEEE/CVF Int. Conf. on Computer Vision
  (ICCV)}, 2023, pp. 10\,895--10\,904.

\bibitem{peng2019mcp}
X.~B. Peng, M.~Chang \emph{et~al.}, ``Mcp: Learning composable hierarchical
  control with multiplicative compositional policies,'' \emph{Advances in
  Neural Information Processing Systems (NeurIPS)}, vol.~32, 2019.

\bibitem{humphreys2024bio}
J.~Humphreys and C.~Zhou, ``Learning to adapt through bio-inspired gait
  strategies for versatile quadruped locomotion,'' \emph{arXiv preprint
  arXiv:2412.09440}, 2024.

\bibitem{escontrela2022adversarial}
A.~Escontrela, X.~B. Peng \emph{et~al.}, ``Adversarial motion priors make good
  substitutes for complex reward functions,'' in \emph{Proc. IEEE/RSJ Int.
  Conf. on Intelligent Robots and Systems (IROS)}.\hskip 1em plus 0.5em minus
  0.4em\relax IEEE, 2022, pp. 25--32.

\bibitem{ampforhumanoid}
Q.~Zhang, P.~Cui \emph{et~al.}, ``Whole-body humanoid robot locomotion with
  human reference,'' in \emph{IEEE/RSJ International Conference on Intelligent
  Robots and Systems}, 2024, pp. 11\,225--11\,231.

\bibitem{peng2024learning}
T.~Peng, L.~Bao \emph{et~al.}, ``Learning bipedal walking on a quadruped robot
  via adversarial motion priors,'' in \emph{Proc. Annual Conference Towards
  Autonomous Robotic Systems (TAROS)}, 2024, pp. 118--129.

\bibitem{2020_Xie_firstsim2real_}
Z.~Xie, P.~Clary \emph{et~al.}, ``Learning locomotion skills for cassie:
  Iterative design and sim-to-real,'' in \emph{Proc. Conf. on Robot Learning
  (CoRL)}, 2020, pp. 317--329.

\bibitem{2021_siekmann_sim2real_nonreference_perodicreward_DRL_e2e_LSTM_PPO_cassie}
J.~Siekmann, Y.~Godse \emph{et~al.}, ``Sim-to-real learning of all common
  bipedal gaits via periodic reward composition,'' in \emph{Proc. IEEE Int.
  Conf. on Robotics and Automation (ICRA)}, 2021, pp. 7309--7315.

\bibitem{2021_UCB_hybridrobotics_sim2real_referencebased_HZD_gaitlibrary_e2epolicy_drl_Cassie_lowpassfilter}
Z.~Li, X.~Cheng \emph{et~al.}, ``Reinforcement learning for robust
  parameterized locomotion control of bipedal robots,'' in \emph{Proc. IEEE
  Int. Conf. on Robotics and Automation (ICRA)}, 2021, pp. 2811--2817.

\bibitem{hwangbo2019learning}
J.~Hwangbo, J.~Lee \emph{et~al.}, ``Learning agile and dynamic motor skills for
  legged robots,'' \emph{Science Robotics}, vol.~4, no.~26, p. eaau5872, 2019.

\bibitem{2023_vanmarum_visionDRL_studentteacher_irregularterrain_PPO_periodicrewardfunction}
B.~van Marum, M.~Sabatelli, and H.~Kasaei, ``Learning perceptive bipedal
  locomotion over irregular terrain,'' \emph{arXiv preprint arXiv:2304.07236},
  2023.

\bibitem{2018_wenhao_yu_DRL_withoutpredefine_symmetrygait_PPO_jointanglePD_}
W.~Yu, G.~Turk, and C.~K. Liu, ``Learning symmetric and low-energy
  locomotion,'' \emph{ACM Transactions on Graphics}, vol.~37, pp. 1--12, 2018.

\bibitem{2020_zhaoming_drl_steppingstones_PPOwithactorcritic_referencefree_simulation}
Z.~Xie, H.~Ling \emph{et~al.}, ``{ALLSTEPS}: Curriculum‐driven learning of
  stepping stone skills,'' \emph{Computer Graphics Forum}, vol.~39, pp.
  213--224, 2020.

\bibitem{2021_siekmann_blind_DRL_stair}
J.~Siekmann, K.~Green \emph{et~al.}, ``Blind bipedal stair traversal via
  sim-to-real reinforcement learning,'' in \emph{Robotics: Science and
  Systems}, 2021.

\bibitem{2022_sim2real_footstepconstraint_OSUDRL_specifytouchdownlocation_actorcritic_PPO_LSTM_plannar_transisionmodel_model-predictiveplanning_CNN_predictnexttdlocation}
H.~Duan, A.~Malik \emph{et~al.}, ``Sim-to-real learning of footstep-constrained
  bipedal dynamic walking,'' in \emph{International Conference on Robotics and
  Automation}, 2022, pp. 10\,428--10\,434.

\bibitem{2022_OSUDRL_steppingstone_referencefree_predictionfeasiblefootsteps_camera_benchmark}
------, ``Learning dynamic bipedal walking across stepping stones,'' in
  \emph{IEEE/RSJ International Conference on Intelligent Robots and Systems},
  2022, pp. 6746--6752.

\bibitem{2023_duan_OSUDRL_heightmap_visionbased_hybrid}
B.~Marum, M.~Sabatelli, and H.~Kasaei, ``Learning vision-based bipedal
  locomotion for challenging terrain,'' \emph{arXiv preprint arXiv:2309.14594},
  2023.

\bibitem{rusu2015policy}
A.~A. Rusu, S.~G. Colmenarejo \emph{et~al.}, ``Policy distillation,''
  \emph{arXiv preprint arXiv:1511.06295}, 2015.

\bibitem{ross2011reduction}
S.~Ross, G.~Gordon, and J.~A. Bagnell, ``A reduction of imitation learning and
  structured prediction to no-regret online learning,'' in \emph{Proc. Int.
  Conf. on Artificial Intelligence and Statistics (AISTATS)}, 2011, pp.
  627--635.

\bibitem{han2024lifelike}
L.~Han, Q.~Zhu \emph{et~al.}, ``Lifelike agility and play in quadrupedal robots
  using reinforcement learning and generative pre-trained models,''
  \emph{Nature Machine Intelligence}, vol.~6, no.~7, pp. 787--798, 2024.

\bibitem{mysore2022multi}
S.~Mysore, G.~Cheng \emph{et~al.}, ``Multi-critic actor learning: Teaching rl
  policies to act with style,'' in \emph{Proc. Int. Conf. on Learning
  Representations (ICLR)}, 2022.

\bibitem{rudin2022learning}
N.~Rudin, D.~Hoeller \emph{et~al.}, ``Learning to walk in minutes using
  massively parallel deep reinforcement learning,'' in \emph{Proc. Conf. on
  Robot Learning (CoRL)}, 2022, pp. 91--100.

\bibitem{peng2018deepmimic}
X.~B. Peng, P.~Abbeel \emph{et~al.}, ``Deepmimic: Example-guided deep
  reinforcement learning of physics-based character skills,'' \emph{ACM
  Transactions On Graphics (TOG)}, vol.~37, no.~4, pp. 1--14, 2018.

\bibitem{andrychowicz2020learning}
M.~Andrychowicz, B.~Baker \emph{et~al.}, ``Learning dexterous in-hand
  manipulation,'' in \emph{The International Journal of Robotics Research},
  vol.~39, no.~1.\hskip 1em plus 0.5em minus 0.4em\relax SAGE Publications,
  2020, pp. 3--20.

\bibitem{hochreiter1997lstm}
S.~Hochreiter and J.~Schmidhuber, ``Long short-term memory,'' \emph{Neural
  Computation}, vol.~9, no.~8, pp. 1735--1780, 1997.

\bibitem{zajac2002biomechanics}
F.~E. Zajac, R.~R. Neptune, and S.~A. Kautz, ``Biomechanics and muscle
  coordination of human walking: Part i: Introduction to concepts, power
  transfer, dynamics and simulation,'' \emph{Gait \& Posture}, vol.~16, no.~3,
  pp. 215--232, 2002.

\bibitem{minetti2000transition}
A.~E. Minetti and R.~M. Alexander, ``Translating resistive force theory to
  human gait transition: A non-linear optimization approach,'' \emph{Journal of
  Experimental Biology}, vol. 203, pp. 2099--2108, 2000.

\bibitem{hubel2013transition}
T.~Y. Hubel and J.~R. Usherwood, ``Transitions from walking to running: New
  insights from a single-body-center-of-mass model,'' \emph{Journal of The
  Royal Society Interface}, vol.~10, no.~88, p. 20120977, 2013.

\bibitem{lee2012centroidal}
J.~Lee and A.~Goswami, ``Centroidal dynamics of a humanoid robot,''
  \emph{Autonomous Robots}, vol.~33, no.~3, pp. 291--311, 2012.

\bibitem{schulman2017proximal}
J.~Schulman, F.~Wolski \emph{et~al.}, ``Proximal policy optimization
  algorithms,'' \emph{arXiv preprint arXiv:1707.06347}, 2017.

\end{thebibliography}
\appendix
\section*{Appendix A: Implementation Details}

\subsection*{A.1 Policy Architecture and Training Parameters}

Our policy follows a recurrent actor-critic architecture with a single-layer LSTM module. The detailed configuration used for training is summarized in Table~\ref{tab:hyperparams}.

\begin{table}[h]
\centering
\caption{Policy Architecture and PPO Training Hyperparameters}
\label{tab:hyperparams}
\begin{tabular}{ll}
\toprule
\textbf{Component}               & \textbf{Setting} \\
\midrule
Policy architecture             & LSTM + MLP \\
Actor hidden dimensions         & [32] \\
Critic hidden dimensions        & [32] \\
RNN type                        & LSTM \\
RNN hidden size                 & 64 \\
RNN layers                      & 1 \\
Activation function             & ELU \\
Initial action noise std        & 0.8 \\
Entropy coefficient             & 0.01 \\
PPO iterations                  & 30,000 \\
\bottomrule
\end{tabular}
\end{table}
\vspace{1em}
All training was conducted using the \texttt{rsl\_rl} framework with a time step of 0.02s and a command update frequency of 10Hz.
\subsection*{A.2 Reward Function Weights}

We apply a gait-conditioned reward routing mechanism, where different subsets of reward terms are activated depending on the current gait mode. The primary reward weights used during training are listed in Table~\ref{tab:rewardweights}.

\begin{table}[h]
\centering
\caption{Active reward terms and weights in the final multi-gait setting.}
\label{tab:rewardweights}
\setlength{\tabcolsep}{6pt}
\renewcommand{\arraystretch}{1.05}
\begin{tabular}{llr}
\toprule
\textbf{Category} & \textbf{Reward Term} & \textbf{Weight} \\
\midrule
\multirow{17}{*}{Regulation rewards} 
& Vertical lin.\ vel.\ penalty ($v_z$) & $-2.0$ \\
& Horizontal ang.\ vel.\ penalty ($\omega_{xy}$) & $-0.05$ \\
& Orientation deviation & $-1.0$ \\
& Base height penalty & $-10.0$ \\
& DOF acceleration penalty & $-2.5\times10^{-7}$ \\
& DOF velocity penalty & $-1\times10^{-3}$ \\
& Collision penalty & $-10.0$ \\
& Action rate penalty & $-0.01$ \\
& DOF limit penalty & $-5.0$ \\
& Alive bonus & $0.15$ \\
& Hip pos.\ deviation & $-1.0$ \\
& Min.\ torso ang.\ vel. & $2.0$ \\
& Waist pitch deviation & $1.0$ \\
& Waist roll deviation & $1.0$ \\
& Waist yaw deviation & $1.2$ \\
& Torso yaw smoothness & $0.8$ \\
& Shoulder roll control & $3.0$ \\
\midrule
\multirow{5}{*}{Arm swing} 
& Arm--leg momentum balance & $5.0$ \\
& Human-like arm swing energy & $0.3$ \\
& Elbow phase tracking & $2.5$ \\
& Arm swing symmetry & $2.0$ \\
& Arm swing--leg amp.\ match & $1.0$ \\
\midrule
\multirow{4}{*}{Walking} 
& Feet swing height penalty & $-15.0$ \\
& Contact & $1.0$ \\
& Straight knee & $0.1$ \\
& Feet drag penalty & $-0.5$ \\
\midrule
\multirow{2}{*}{Standing} 
& Contact (standing) & $2.5$ \\
& Base motion (standing) & $2.5$ \\
& Pose (soft upper) & $4.0$ \\
& Feet alignment & $0.5$ \\
& Uprightness & $1.0$ \\
& Feet drag penalty & $-0.2$ \\
& Feet flatness & $2.5$ \\
& Stillness bonus & $2.0$ \\
\midrule
\multirow{4}{*}{Walk→Stand (W2S)} 
& Feet swing height penalty & $-20.0$ \\
& Contact & $1.0$ \\
& Smooth slowdown & $0.1$ \\
& Feet drag penalty & $-0.1$ \\
\midrule
\multirow{6}{*}{Running} 
& Contact & $1.0$ \\
& Feet drag penalty & $-1.0$ \\
& Short ground contact & $0.2$ \\
& Feet swing height penalty & $-25.0$ \\
& Push-off reward & $1.0$ \\
& Forward velocity reward & $0.2$ \\
\midrule
\multirow{5}{*}{Run→Walk (R2W)} 
& Contact & $1.0$ \\
& Smooth slowdown & $1.0$ \\
& Transition to walk speed & $0.5$ \\
& Feet swing height penalty & $-25.0$ \\
\midrule
\multirow{2}{*}{Task rewards} 
& Tracking lin.\ vel. & $1.5$ \\
& Tracking ang.\ vel. & $1.0$ \\
\bottomrule
\end{tabular}
\end{table}

\end{document}